\documentclass[runningheads]{llncs}
\usepackage[T1]{fontenc}
\usepackage{graphicx}
\begin{document}
\title{Damage Classification for 3D Point Cloud Data via 3D Data Analysis and Vision Foundation Model-based 2D Projections}
%
%
\author{Evan Perez\inst{1} \and
Kalelo Dukuray\inst{1} \and
Erika Ardiles-Cruz\inst{2} and Jie Wei\inst{1*}}
\authorrunning{Perez, Dukuray, Ardiles-Cruz, Wei}
%
\institute{Dept. of Computer Science, City College of New York, NY 10031, USA \and
Air Force Research Lab, Rome, NY 13441, USA\\
\email{jwei@ccny.cuny.edu}  \footnote{\dag~This work is publicly released with No. AFRL-2026-3023}\\
}
\maketitle              
\begin{abstract}

Fine-grained damage classification of 3D point cloud data (PCD) remains a persistent challenge, constrained by high computational demands and limited labeled data. This study examines two methods: 1) {\it 3D PCD-based damage assessment (3PDA)} algorithm: the damage analysis is derived from 3D PCD using PCD segmentation, topological data analysis (TDA), and anomaly detection methods, and 2) {\it 2D projection damage assessment (2PDA)}: the multi-view 2D projection of 3D PCD is used for damage analysis using large vision foundation models (VFMs).

In our 3PDA analysis algorithm, TDA is used to derive compact representations of 3D PCD segmented by pointNet, which are then integrated with anomaly detection algorithms to quantify structural degradation. We show that TDA effectively compresses geometric structure from VFM-segmented components into discriminative feature vectors and that anomaly detection models can reliably distinguish components with varying damage severity using only 3D PCD inputs. In the 2D projection analysis algorithm, we leverage large VFMs for granular damage detection by projecting 3D PCD into 2D views. These projections allow VFM based models to achieve competitive classification performance while requiring only a fraction of the computational cost associated with full 3D data processing. Our results demonstrate that 2D VFM pipelines in 2PDA can perform strongly on fine-grained damage classification tasks, highlighting their viability as lightweight, resource-efficient alternatives to traditional 3PDA architectures.

Comparative evaluation shows that the 3PDA  attains higher accuracy but only for a narrow subset of object geometries and at substantially higher computational cost due to its reliance on TDA and the scarcity of high-fidelity 3D datasets. In contrast, the 2PDA algorithm yields slightly lower accuracy but offers an order of magnitude reduction in time complexity and generalizes across a far broader range of object categories.

\end{abstract}
\section{Introduction}
It is of great importance in civilian and military applications to assess the damage caused by natural disasters \cite{wei2022deep} and on the battlefield effectively for decision making and resource allocation \cite{wei2021nida}.
Multidimensional datasets, such as 3D point cloud data (3D PCD) collected by  Light Detection and Ranging (LiDAR) or photogeometry \cite{thompson2023multi}, offer immense value in storing a large amount of detailed information per sample to describe and characterize patterns for situation analysis, which effectively defies the differing weather, lighting, and event actions. Storing information about a physical object in three-dimensional space is a common and often preferred method of characterizing physical objects, with three-dimensional representations being the forefront of how scientists capture and store physical data. However, with great detail also comes the need for proportional storage and computational cost, which also affects the speed at which a multidimensional dataset can be queried and processed/analyzed as a guide for actions. 

In various practical settings, three-dimensional data can be easily acquired through sensor data, which provides high-resolution three-dimensional models of large objects. With LiDAR data that carries large amounts of essential information for analysis, it can be difficult to replace this methodology with a data collection alternative that reduces storage and computational requirements, especially when data collection methods are established and fixed by regulations and preferences. Dimensionality reduction becomes a favorable post-data collection solution to extracting essential data from large, often noisy datasets collected via LiDAR, while reducing the space and power needed to search and analyze it. 

This study examines two methods: 1) The {\it 3D PCD damage assessment (3PDA)} algorithm: direct direct analysis of 3D PCD using various 3D PCD analysis methods, such as pointNet \cite{qi2017pointnet}, topological data analysis (TDA) \cite{schrader2025towards}, and anomaly detection methods \cite{liu2008isolation}, and 2) The {\it 2D projection damage assessment (2PDA)} algorithm: multi-view 2D projection of 3D data to enable the use of large vision foundation models (VFMs), where 3D damage localization \cite{wei2025effective} on aircraft point clouds cast as a 2D problem amenable to lightweight vision foundation models. 

\begin{figure}
\centering
\includegraphics[width=0.8\textwidth]{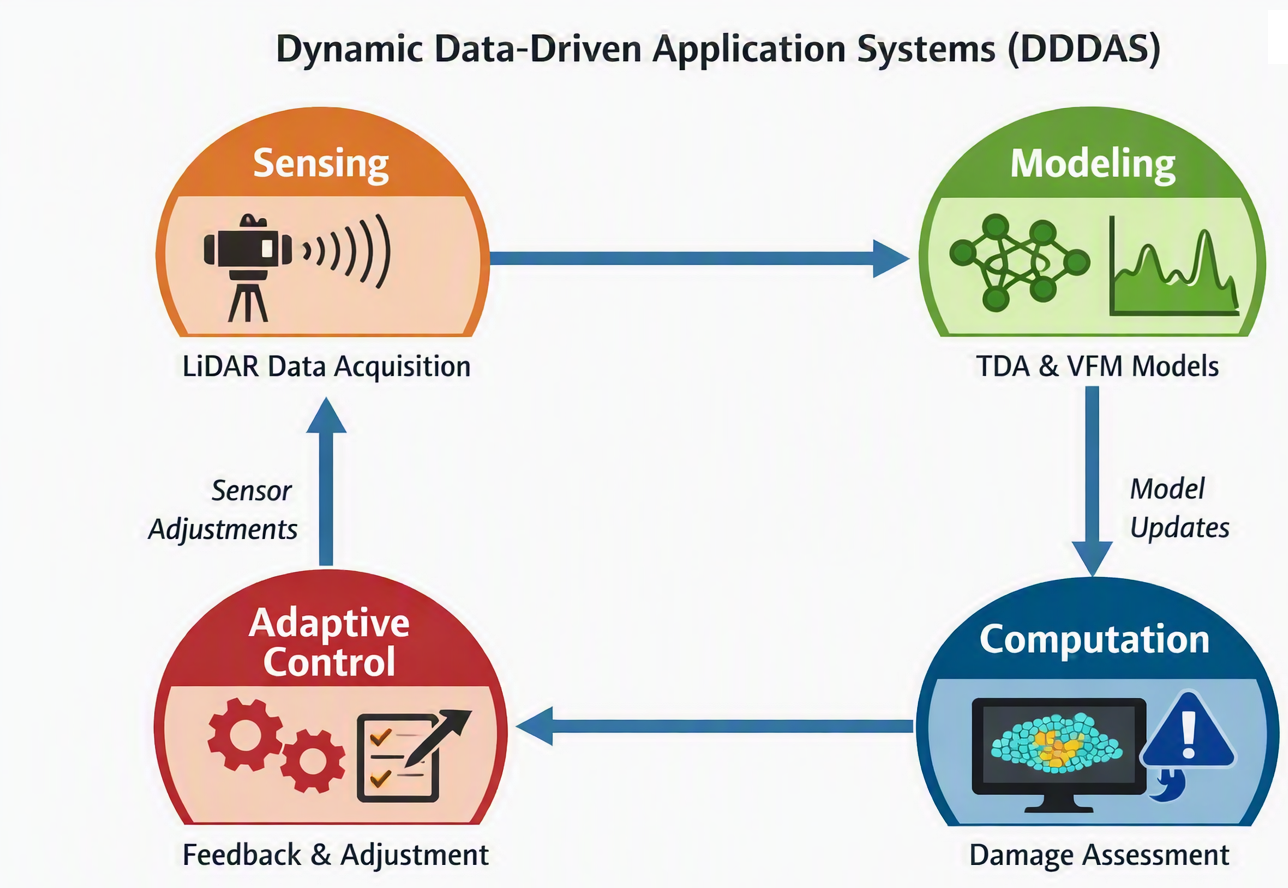}
\caption{Dynamic Data‑Driven Application Systems (DDDAS) framework of this work, where LiDAR sensing, modeling, computation, and adaptive control interact continuously for real‑time aircraft damage assessment and system optimization.} \label{fig0}
\end{figure}

As shown in Fig. \ref{fig0}, the connection of this work to the broader Dynamic Data‑Driven Application Systems (DDDAS) paradigm arises from the way both 3PDA and 2PDA pipelines adapt computation to streaming or situational data conditions. Multidimensional datasets, such as 3D point cloud data collected by LiDAR, offer immense value in storing a large amount of detailed information per sample to describe and characterize patterns for situation analysis. In a DDDAS framework, sensing and computation are tightly coupled: models dynamically refine their analysis based on incoming data, and sensing strategies can be adjusted based on computational feedback. The proposed 3PDA pipeline exemplifies this by using TDA‑derived structural descriptors and anomaly detection to update damage assessments as new PCD becomes available, while the 2PDA pipeline leverages lightweight VFM‑based projections \cite{yuan2016novel} to provide rapid, resource‑efficient assessments that can be invoked adaptively depending on mission constraints. Together, these methods illustrate how dynamic integration of sensing, modeling, and computation—core principles of DDDAS—can enhance real‑time damage assessment for aircraft and other critical assets.

\section{3PDA: Direct Damage Assessment in 3D PCD}

\subsection{Localized Aircraft Damage Assessment}
Automating the process of characterizing large aircraft 3D PCD is a challenge that has been one of the main focal points of our research. In particular, finding a solution that automates the identification of damaged aircraft parts is an area of great interest, as healthy aircraft LiDAR scans are abundant, but damaged aircraft scans are extremely scarce. Machine learning solutions are turned to in order to find complex patterns in LiDAR point cloud data, but processing times are long, and large amounts of both healthy and damaged aircraft LiDAR samples are required to train models on. This presents a class imbalance issue, where one class type dominates as the majority (healthy aircraft) while another class severely lacks in samples gathered (damaged aircraft), making machine learning solutions difficult to implement without sacrificing training data size through under-sampling. 

\subsection{Methodology}
	We developed the 3PDA algorithm, which combines the power of neural networks to ingest and classify raw PCD, the feature extraction capability of TDA to reduce dimensions while retaining high-level descriptors of aircraft PCD, and the robustness of unsupervised ensemble learning forest algorithms to learn common features of healthy aircraft LiDAR scans and be able to identify anomalous, aka damaged, aircraft PCD. With this approach, raw PCD files can be easily uploaded via a graphical user interface, and a Python-based pipeline will apply trained machine learning algorithms and TDA to the PCD file in order to provide a detailed analysis of the structural integrity status of major parts of the aircraft being ingested. The main parts our system analyzes are the body, wings, and tail of the aircraft, which produce a score in a range between -1 and 1 to describe the damage level of each aircraft part. 

    As seen above, our approach breaks down the problem into smaller parts and, at each stage, achieves a different step towards assessing the damage of the original aircraft file’s parts. The major steps are listed in order as follows:
1) Trained Part Segmentation model identifies Wing, Body, Tail of aircraft 
2) Topological Data Analysis (TDA) procedure is applied to the PCD parts to extract features and reduce dimensions
3) Feature vectors of parts are fed to the anomaly detection algorithm to identify anomalous (damaged) parts 

Each step is explained in detail below.

\noindent 1. {\bf PointNet Segmentation}: For part segmentation, we utilize the robust PointNet deep learning architecture to ingest and train on raw point cloud data to identify patterns between the various, pre-labeled parts of the plane. There are multiple versions of PointNet that achieve different tasks, such as classification or semantic segmentation, but for our research, we rely on the part segmentation variant for our initialization phase of the system. PointNet is particularly useful as it first introduced a method for feeding raw point clouds into a machine learning algorithm without any major preprocessing steps or transformations. Figure 5 shows an example of PointNet’s capabilities from the original paper. This enables our system to take a raw point cloud as input and clearly separate the wing, body, and tail portions of the original point cloud aircraft. We adopt this deep learning architecture only at the first phase of the system, and not throughout the rest of the process, primarily for the segmentation capability, which allows us to focus on analyzing smaller point clouds rather than one large, continuous point cloud file. Not only does this reduce the size of data processed at later steps, but it ensures that the system extracts meaningful data from each distinct plane part separately, rather than running analysis all together where feature descriptors might leak into one another. Fig. \ref{fig2} demonstrates the output for the segmentation phase of the pipeline, which uses an aircraft point cloud as the initial input.

\noindent 2. {\bf Alpha-Complex-based TDA}

After the segmentation process is completed, we now have a set of points that contain the labels for the aircraft parts that they represent out of the whole point cloud. With this, we break the problem down into three distinct point clouds: Wing, Body, and Tail of the aircraft. With these separate point cloud objects, we can proceed to extract their topological features and significantly reduce their size with topological data analysis. 

 In this research experiment, we implement a topological data analysis method called Alpha Complex, which has a worst-case time complexity of O($N^3$), but typically O($N \log(N))$. 
For our machine learning system, Alpha-Complex-based TDA was applied to extract high-level topological features and reduce dimensions from the dataset. Specifically, TDA was applied to a sub-dataset that was made from the original ShapeNet dataset, one that contains three class categories for Wing, Body, and Tail labeled aircraft point clouds. With this process applied, these point cloud objects are now transformed into separate feature matrices, containing valuable topological descriptors about each aircraft part. In particular, we are interested in the numerical descriptors that describe the voids and loops within the object, which can be critical in determining if the aircraft part is damaged or not.

\noindent 3. {\bf Anomaly detection for damage detection}

In the final component of our PCD damage assessment system, we utilize unsupervised ensemble learning to learn common features from our TDA-extracted feature vectors of healthy aircraft objects. We tried out two different anomaly detection methods in this step: Isolation Forest and one-class Support Vector Machine (SVM) \cite{manevitz2001one}.

As the name suggests, Isolation Forest works by isolating data points that have distinct features from a majority class using random partitioning. The algorithm uses the very fact that there are rare and few samples in the dataset that differ significantly from the rest in order to quickly separate them. Randomly selected features and thresholds are used to split the data, which is able to split the “anomalies” from the normal data quite efficiently. Data points that are split a small number of times are considered anomalous, as they do not require that many partitions to separate them from the rest of the normal data. After identifying what makes anomalies different from the majority class, an anomaly score is assigned to the data point to represent how much it differs. Scores range between -1 and 1, with -1 being the most different or “anomalous” from the normal data, and a value closer to 1 meaning that the data is normal. 

One-class SVM creates a decision boundary around the majority class, often the only class that is trained on, and maximizes the margin around those “normal” samples. Like Isolation Forest, the main technique focuses on learning features and patterns from the normal group of data to be able to identify when a new sample of data is different or “anomalous”. It characterizes the anomaly detection output by assigning scores around 0, where positive numbers indicate normal samples and negative numbers indicate anomalies. Unlike an isolation forest, which has a range between -1 and 1, one-class SVM has a wider range of float values using the decision function, so anomaly scores can take on small or large values. With the decision function in the Scikit-Learn implementation of one-class SVM, this is the distance from the hyperplane made by the model.

\subsection{Results}
Experimental results using the 3PDA algorithm described in the prior subsection are demonstrated here to shed light on the efficacy of this work.

\begin{figure}
\centering
\begin{tabular}{cc}
\includegraphics[width=0.70\textwidth]{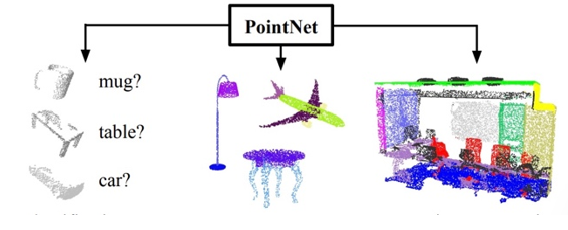} &
\includegraphics[width=0.30\textwidth]{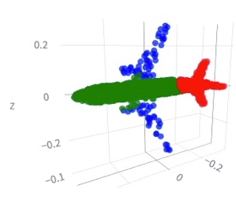}\\
(a) & (b)
\end{tabular}
\caption{PointNet-based PCD segmentation. (a) Use of PointNet for object classification. (b) PointNet separates the initial aircraft 3D PCD into its respective Wing (blue), Body (green), and Tail (red) segments.} \label{fig1}
\end{figure}

\begin{figure}
\centering
\includegraphics[width=0.85\textwidth]{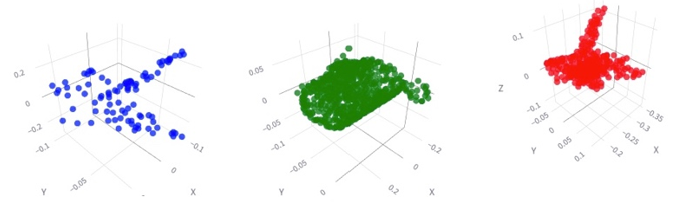}\\
(1)  \hspace*{3cm} (2)  \hspace*{3cm} (3)
\caption{Segment-based damage assessment output of the anomaly detection systems, which shows three different aircraft parts evaluated with part-specific anomaly detection models. From left to right: 1) Wing, score: -0.119, moderate damage; 2)Body, score: 0.000, healthy; 3)Tail, score: 0.000, healthy.} \label{fig2}
\end{figure}

Fig. \ref{fig1} demonstrates the results of the 3D PCD local parts segmentation generated by pointNet. The General utilities of pointNet are shown in Fig. \ref{fig1}a, while the three local parts of an aircraft PCD generated by pointNet are depicted in Fig. \ref{fig1}b, which sets the stage for the TDA-based anomaly detection for damage assessment. In Fig. \ref{fig2}, the part-specific damage assessment called out by the trained Isolation Forest is illustrated, which conforms to human expert evaluations. 

Despite being a potentially promising method for anomaly detection with feature vectors of point cloud data, our anomaly detection experiments using one-class SVM turned out to flag many false alarms, and even crank out much larger negative scores for anomalies on true healthy parts compared to the true damaged part, which was not correctly found as an outlier. In this work, our default anomaly detection method for damage assessment is thus set to be Isolation Forest.

Due to the immense data size and negative impacts of outliers in TDA evaluations, in our tests, we used random sampling to inspect the consequences of reducing the data size and outliers in our damage assessment work. After testing our many different choices, such as Farthest Point Sampling, we discovered that 
regular random sampling proved to be superior, with minimal false alarms and accurate true positive anomaly detection on aircraft parts with sparse points that simulated damage. In testing this approach, our system accurately assigns scores of 0.0 to the body and tail, which are healthy, and then assigns our simulated damaged wing a score of -0.19, which is an accurate score for our moderate damage simulation. Farthest Point Sampling, on the other hand, assigns the healthy body to a severe damage score of -0.275, the healthy tail a minimal damage score of -0.066, and the simulated damaged wing a minimal damage score of -0.049. 
Random sampling perhaps works better in this case due to a higher likelihood of reducing the number of outlier points sampled, and thus ensuring the TDA process did not include artifact descriptors that would have negatively affected the vector matrix output. 

\section{2PDA algorithm: damage assessment from 2D Projections}
This section details the 3PDA algorithm to assess 3D PCD damage by projecting 3D PCD to 2D images.

\subsection{Methodology}
The dominant pathway for damage assessment on 3D PCD: running deep 3D neural networks such as PointNet directly on the cloud is computationally demanding and data-hungry, particularly when the task moves from whole-component classification to fine-grained, spatially localized damage detection, as detailed in the preceding section. In view of this problem, we seek to choose a different route for damage assessment that can use 2D projections from a 3D PCD to provide part-specific damage levels. Vision Foundation models (VFMs) such as YOLO \cite{varghese2024yolov8}, which learn lots of knowledge from their pre-trained vision data, can be fine-tuned using our annotated 2D damage images to effectively assist us in assessing damage.

In this work, the 3D PCD is projected onto the three orthogonal planes, XY, XZ, and YZ, using a fixed visual style (Viridis colormap on a black background with a fixed marker size) matched to the distribution on which the models were trained. On each of the three views, two YOLO models are run in parallel: a parts detector that localizes wings, body, and tail, and a damage detector that localizes damaged regions. Overlapping bounding boxes from the two detectors are matched with an IoU threshold, producing a per-view mapping from each damage region to the specific component or components it affects.

\begin{figure}
\begin{center}
\begin{tabular}{cc}
\includegraphics[width=0.35\textwidth]{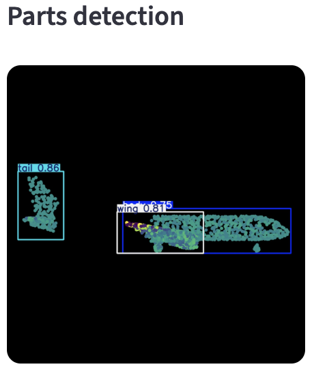} & \hspace*{1cm}
\includegraphics[width=0.35\textwidth]{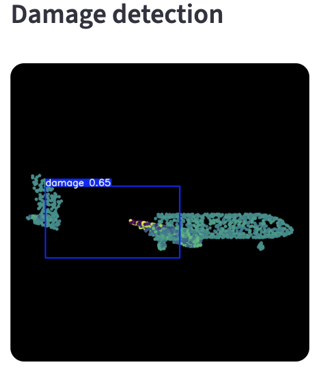}
\end{tabular}
\end{center}
\caption{Damaged aircraft parts detection by 2PDA algorithm. The parts detector (left) localizes parts bounding boxes; the damage detector (right) localizes the damaged regions.} \label{fig3}
\end{figure}

The output is component-attributed, spatially localized damage assessment delivered by a pair of lightweight 2D detectors operating on a small group of orthogonal projections, rather than a single heavy 3D pipeline. Relative to the PointNet-style 3D baseline used elsewhere in the broader application, this approach maintains discriminative performance on the harder localization task while bringing the compute and data requirements in line with standard 2D detection workflows. The projection-based reframing therefore extends cleanly from the earlier whole-component classification work to fine-grained spatial localization, which supports the broader claim that small ensembles of 2D vision-foundation-model inferences can substitute for dedicated 3D models at a small fraction of the compute. Figure 1 shows the segment’s interface analyzing synthetic damage on the XY projection of an aircraft point cloud.

\subsection{Experimental Results}

To gain insights into the performance of this 2D projection-based damage assessment, synthetic damage is introduced to healthy aircraft point clouds by removing a spherical chunk of points at a random location, with severity controlled by the radius of the removed region. Fig. \ref{fig3} shows the segment’s interface analyzing synthetic damage on the XY projection of an aircraft point cloud.



\section{Conclusion}
 To effectively assess damages from 3D PCD, we developed two different methods: 1) The 3D PCD-based damage assessment (3PDA) algorithm: we first call pointNet for part segmentation, then apply a topological data analysis procedure to obtain data features, and then classify the damage levels of each component. 2) 2D projection damage assessment (2PDA) algorithm: We first project the 3D PCD to 2D images, then the fine-tuned Vision foundation models are used to produce damage levels from these 2D projections, where the immense knowledge learned from pre-trained data can be effectively transferred. Preliminary results of both approaches show encouraging performance. A methodology-level fusion will be explored next to fuse these two orthogonal methods to achieve more optimized damage assessment performance.

\subsubsection{Acknowledgements} The authors would like to gratefully acknowledge the support of ML-RCP and AFRL FA950-21-1-0082.

%
%
%
\bibliographystyle{splncs04}
\bibliography{mybib}

\end{document}